# Illustrating Color Evolution and Color Blindness by the Decoding Model of Color Vision

LU, CHENGUANG

Survival99 at gmail.com, http://survivor99.com/lcg/english

A symmetrical model of color vision, the decoding model as a new version of zone model, was introduced. The model adopts new continuous-valued logic and works in a way very similar to the way a 3-8 decoder in a numerical circuit works. By the decoding model, Young and Helmholtz's tri-pigment theory and Hering's opponent theory are unified more naturally; opponent process, color evolution, and color blindness are illustrated more concisely. According to the decoding model, we can obtain a transform from RGB system to HSV system, which is formally identical to the popular transform for computer graphics provided by Smith (1978). Advantages, problems, and physiological tests of the decoding model are also discussed.

## 1. Introduction

Young and Helmholtz's tri-pigment theory and Hering's opponent theory on color vision have been competing for a long time. A compromising viewpoint accepted widely is that color signals exist in tri-pigments at the zone of visual cones and in opponent signals at the zone of visual nerves (De Monasterio and others, 1975). The mathematical model with this viewpoint is the zone model (Judd, 1949). There are many improved versions (Hurvich etc., 1957; Walraven, 1961; Hunt, 1982). Yet, why are color signals processed in this way and how has color vision been evolving? The answers are still unclear. To answer these questions, I built a model of color vision named the decoding model (Lu, 1986), which is new version of zone model, and verified it by predicting color appearance (Lu, 1989). Recently, I found that a popular transform from RGB system to HSV systems for computer graphics (A. R. Smith,1978) is formally identical to the transform based on the decoding model. This means that the decoding model is also practical. This paper is to introduce the decoding model and the transform, and to explain, opponent process, color evolution, and color blindness pictorially.

## 2. Fuzzy 3-8 Decoding

The binary 3-8 decoder is frequently used in computers or numerical circuits for selecting one register or memory from eight. If $B$, $G$, and $R$ are binary switching variables, i.e. $B$, $G$, and $R$ take values in the set $\{0,1\}$, as three inputs to a 3-8 decoder, then eight outputs will be $[\bar{B}\bar{G}\bar{R}]$, $[\bar{B}\bar{G}R]$, $[\bar{B}G\bar{R}]$, $[\bar{B}GR]$, $[B\bar{G}\bar{R}]$, $[B\bar{G}R]$, $[BG\bar{R}]$, and $[BGR]$ ([...] Denotes a logical expression).

For example, if $B=G=0$ and $R=1$, then $[\bar{B}\bar{G}R]=1$, otherwise $[\bar{B}\bar{G}R]=0$.

Let $B$, $G$, and $R$ represent the outputs of three cones and a color be denoted by a vector ($B$, $G$, $R$). Hence $[\bar{B}\bar{G}\bar{R}]$, $[\bar{B}\bar{G}R]$, …, $[BGR]$ stand for the magnitude of eight color signals: blackness, redness, ..., whiteness (see Table 1).

**Table 1: Relation between *B*, *G*, and *R* and values of eight output codes or color signals**

| B G R | Blackness $[\overline{B}\overline{G}\overline{R}]$ | Redness $[\overline{B}\overline{G}R]$ | Yellowness $[\overline{B}GR]$ | Greenness $[\overline{B}G\overline{R}]$ | Cyanness $[BG\overline{R}]$ | Blueness $[B\overline{G}\overline{R}]$ | Magentaness $[B\overline{G}R]$ | Whiteness $[BGR]$ |
|---|---|---|---|---|---|---|---|---|
| 0 0 0 | 1 | 0 | 0 | 0 | 0 | 0 | 0 | 0 |
| 0 0 1 | 0 | 1 | 0 | 0 | 0 | 0 | 0 | 0 |
| 0 1 1 | 0 | 0 | 1 | 0 | 0 | 0 | 0 | 0 |
| 0 1 0 | 0 | 0 | 0 | 1 | 0 | 0 | 0 | 0 |
| 1 1 0 | 0 | 0 | 0 | 0 | 1 | 0 | 0 | 0 |
| 1 0 0 | 0 | 0 | 0 | 0 | 0 | 1 | 0 | 0 |
| 1 0 1 | 0 | 0 | 0 | 0 | 0 | 0 | 1 | 0 |
| 1 1 1 | 0 | 0 | 0 | 0 | 0 | 0 | 0 | 1 |

Now, suppose that *B*, *G*, and *R* vary from the binary switching variables into the continuous switching variables, i.e. *B*, *G*, and *R* take continuous values in set ［0, 1］. With the special continuous-valued logic or fuzzy logic (Lu, 1991), we can extend the binary 3-8 decoding into the fuzzy 3-8 decoding (Lu, 1986). The values of output codes are illustrated by Figure 1.

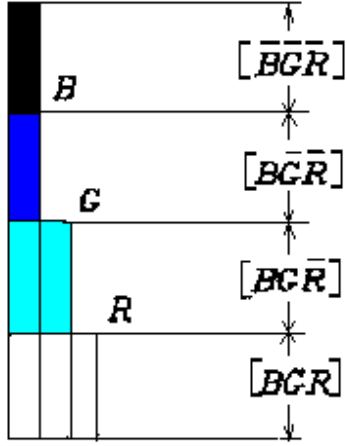

Figure 1: Relation between three input codes *R*, *G*, and B and eight output codes of the fuzzy 3-8 decoder (When *B*>*G*>*R*, the values of four output codes are shown and other values are equal to zero)

Let max(*a*, *b*) stand for the maximum of *a* and *b*, min(*a*,*b*) for the minimum of *a* and *b*, and so on. Hence

$$[\overline{B}\overline{G}\overline{R}] = 1 - \max(B, G, R)$$
$$[\overline{B}\overline{G}R] = \max(0, R - \max(B, G))$$
$$[\overline{B}GR] = \max(0, \min(G, R) - B)$$
$$[BGR] = \min(B, G, R)$$

（1）

The others can be calculated in similar ways.

## 3. Transform from RGB system to HSV System

Assume *B*, *G*, and *R* are tri-stimulus valves from cones. How do we simulate the visual system to obtain *H* (hue), *S* (saturation), and *V* (brightness) from *B, G,* and *R*?

For any given color denoted by (*B, G, R*), there is an equation

$$(B,G,R) = [\overline{BG}R](0,0,1) + [\overline{B}GR](0,1,1) + [\overline{B}G\overline{R}](0,1,0) + [BG\overline{R}](1,1,0) \\ + [B\overline{GR}](1,0,0) + ([\overline{B}G\overline{R}](1,0,1) + [BGR](1,1,1) \quad (2)$$

which means that any color can be decomposed into the combination of white and six unique colors in different ratios. In the equation, (0, 0, 1) stands for the most saturated red, i.e. unique red, and the coefficient $[\overline{BG}R]$ is the redness of the color (*B, G, R*), and so on.

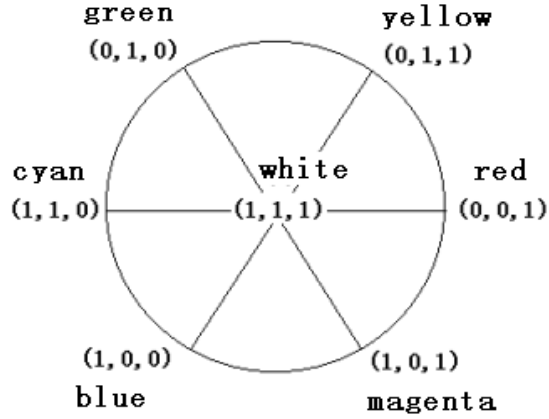

Figure 2. Decomposition of color (*B, G, R*)

It is coincident that only three items on the right of equation (2) may be non-zero for a given color and the three cardinal vectors or unique colors must be at the three vertexes of one of six sectors in Figure 2. Hence equation (2) can be changed into

$$(B,G,R) = m_1 e_1 + m_2 e_2 + [BGR](1,1,1) \quad (3)$$

where $e_1$, $e_2$ are two cardinal vectors or unique colors and $m_1$, $m_2$ are corresponding coefficients or magnitude of output codes.

Suppose the angles at which $e_1$ and $e_2$ are located (see Figure 2) are $H_1$ and $H_2$. Let

$$H = (m_1 H_1 + m_2 H_2)/(m_1 + m_2) \\ C = m_1 + m_2 \\ V = m_1 + m_2 + [BGR] = \max(B,G,R) \\ S = C/V \quad (4)$$

Then *H, C, V, S* will represent hue, colorfulness, brightness, and saturation of (*B, G, R*) properly if *B, G* and *R* are obtained from appropriate linear and nonlinear transforms of spectral tri-stimulus values *X, Y, Z* (Lu, 1989). According to the decoding model, the relation between brightness, colorfulness, whiteness, blackness and *B, G* and *R* is shown in Figure 3, where med(*B, G, R*) is the medium one or second one of *B, G* and *R*. For example, med(1,3,5)=3, med(1,2,5)=2, med(1,5,5)=5, med(1,1,5)=1.

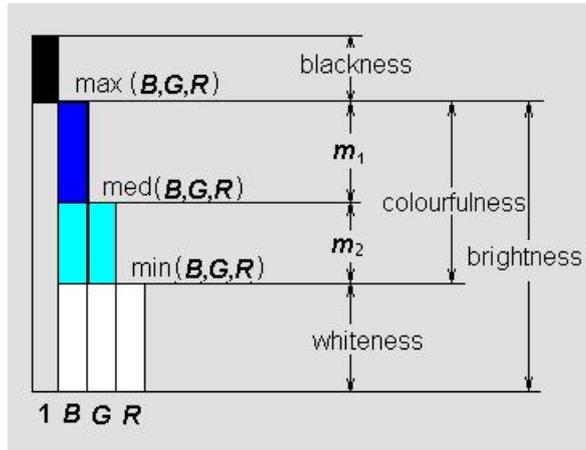

Figure 3 Relation between *B*, *G*, *R* and brightness, colorfulness, whiteness, and blackness

Recently, I found the above transform has been proposed earlier by A. R. Smith (1978), and introduced by many scholars for computer graphics (J.D. Foley, A. Van Dam, 1984). The detailed programs can be seen on web pages[1] This shows that the decoding model is also practical. The different is that 1)*B, G,* and *R* in A. R. Smith's transform are the magnitudes of signals of primary colors to stimulate a pixel of CRT instead of tri-stimulus values from visual cones; 2)A. R. Smith used "if-then" programming language rather than logical operations; 3)The following opponent process only exists in the decoding model.

## 4. Opponent Process

We use Venn's Diagram to show the essence of the process. Let $\cap, \cup, ^c$ denote the three set operations: intersection, union, and complement respectively; *B, G,* and *R* represent the three circular fields respectively (see Figure 4). For convenience, we also use "¯" for complement operation and omit $\cap$. Then, the eight fields can be represented by $[\overline{B}\overline{G}\overline{R}]$, $[\overline{B}\overline{G}R]$, $[\overline{B}G\overline{R}]$, $[\overline{B}GR]$, $[B\overline{G}\overline{R}]$, $[B\overline{G}R]$, $[B G\overline{R}]$, and $[BGR]$.

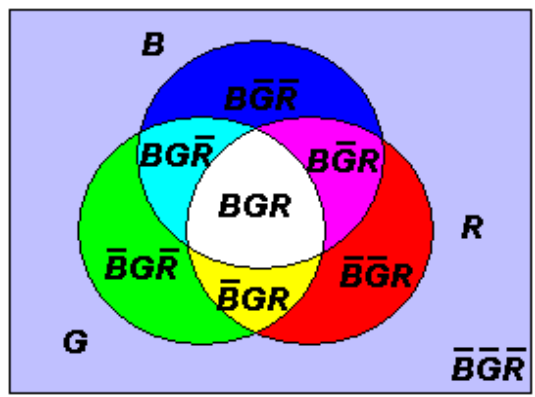

Figure 4. Venn's diagram showing the logic of the opponent process

---

[1] http://www.acm.org/jgt/papers/SmithLyons96/hsv_rgb.html

From *B, G* and *R*, we can first get
$$M = BG \cup BR \cup GR \tag{5}$$
which represents the trefoil (the intersecting fields of two or three of *B, G*, and *R*). Then, we have
$$B\overline{M} = B\overline{(BG \cup BR \cup GR)} = B(\overline{B} \cup \overline{G})(\overline{B} \cup \overline{R})(\overline{G} \cup \overline{R})$$
$$= B(\overline{B} \cup \overline{BG} \cup \overline{BR} \cup \overline{GR}) = B\overline{G}\overline{R} \quad \text{(blue area)} \tag{6}$$
where DeMorgan Law is used. Similarly, there are
$$\overline{B}M = \overline{B}GR \quad \text{(yellow area)} \tag{7}$$
$$G\overline{M} = \overline{B}G\overline{R} \quad \text{(green area)} \tag{8}$$
$$\overline{G}M = B\overline{G}R \quad \text{(magenta area)} \tag{9}$$
$$R\overline{M} = \overline{B}\overline{G}R \quad \text{(red area)} \tag{10}$$
$$\overline{R}M = BG\overline{R} \quad \text{(cyan area)} \tag{11}$$

Now let *B, G* and *R* denote three receptor outputs and take values in the set [0, 1], and the set operations be replaced by the fuzzy logic operations: ∨, ∧, ¯ (∨ stands for maximum, ∧ for minimum and ¯ can be omitted). First we obtain the medium one of *B, G,* and *R* (see Figure 5):
$$M = med(B,G,R) = [BG \vee BR \vee GR]$$
$$= \max(\min(B,G), \min(B,R), \min(G,R)) \tag{12}$$

Then we get three opponent signals, blueness-yellowness ($M_{BY}$), greenness-magentaness ($M_{GM}$), and redness-cyanness ($M_{RC}$). The calculation as follows are surprisingly simple:

$$M_{BY} = B - M = \begin{cases} +[B\overline{G}\overline{R}], & B \geq M, \\ -[\overline{B}GR], & B < M; \end{cases} \tag{13}$$

$$M_{GM} = G - M = \begin{cases} +[\overline{B}G\overline{R}], & G \geq M, \\ -[B\overline{G}R], & G < M; \end{cases} \tag{14}$$

$$M_{RC} = R - M = \begin{cases} +[\overline{B}\overline{G}R], & R \geq M, \\ -[BG\overline{R}], & R < M; \end{cases} \tag{15}$$

The opponent process corresponding to different monochromatic lights is shown in Figure 5, where for convenience the three response curves are assumed. We can also consider the left-upper part of Figure 5 as a Venn's diagram. There are eight divided fields. The length of the part of a vertical line on a field is just the magnitude of the corresponding unique color signal. These fields can illustrate the change of color perception caused by different monochromatic lights well.

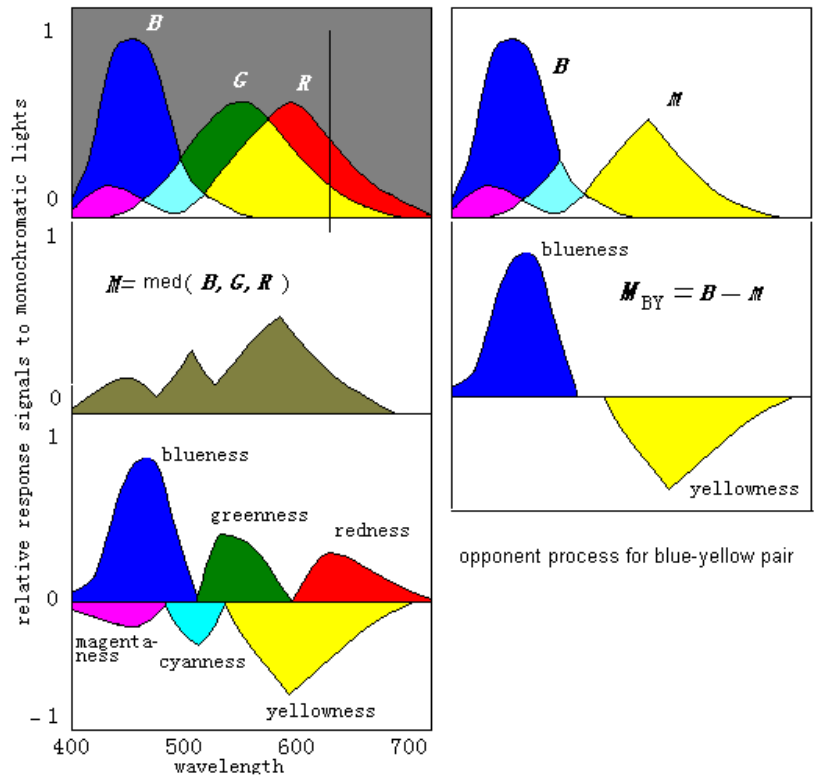

Figure 5: Opponent process corresponding to different monochromatic lights

## 5. Physical Model

The diagram of the principle of the opponent process in the decoding model is shown in Figure 6.

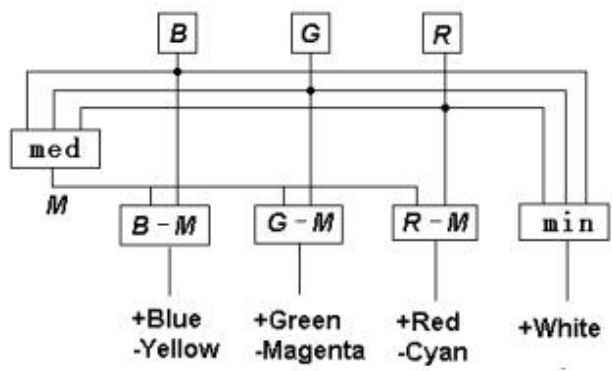

Figure 6. Symmetrically opponent process in the decoding model

In order to demonstrate the process of color signals both in the retina and in the cortex, I built a completely physical model of color vision (see Figure 7), which works as well as I had expected.

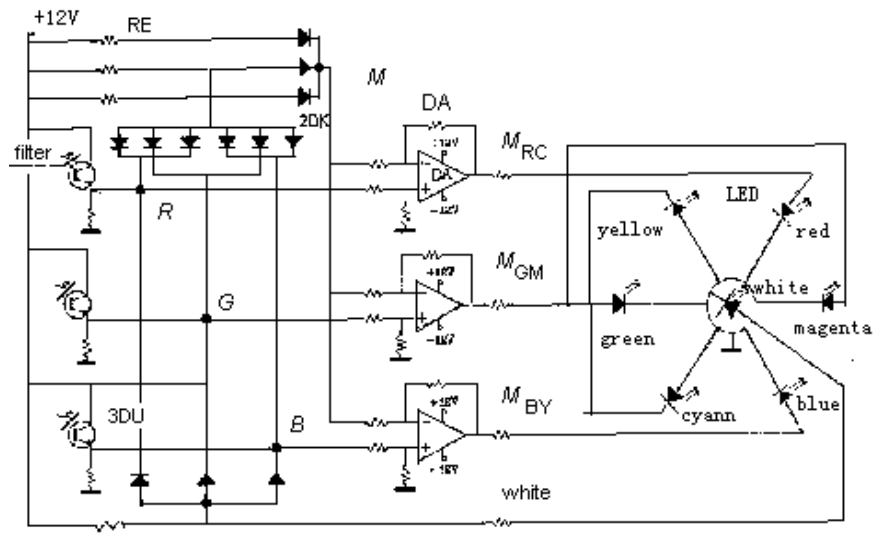

Figure 7. Diagram of principle of the physical model of color vision

Here 3DU is a phototransistor, which imitates a cone; DA is a differential amplifier, which imitates a bipolar; LED is a light emitting diode, which is assumed to be color cells in cortex; RE is a resister and 2DK is a diode. The array of diodes and resisters on the upper left is assumed to be a horizontal cell and provides output $M=\text{med}(B,G,R)$.

The physical model suggests that, in the cortex, there be seven color cells, which receive white and six unique color signals; the brain produce brightness and colorfulness by simple addition, and turns out hue and saturation by the weighing method. Perhaps there are also some processes of color signals on lower levels in retina. For example, some white cells probably receive faster conducting signals directly from cones and rods (Kaplan, 1982). The decoding model does not cover this subject. Thus, it does not provide a measure as luminance $Y$ in CIE XYZ or light value $V$ in the Munsell system, but $V$ for brightness. The process of spatial information is also not considered in the decoding model.

## 6. The Evolution of Color Vision

According to the decoding model, we can easily explain the evolution of color vision by splitting sensitivity curves of visual cones (see Figure 8). Please imagine that curves $R(\lambda)$ and $G(\lambda)$ gradually approach one another to become one curve named $Y(\lambda)$. Then we would see the fields representing red, green, cyan, and magenta disappear gradually. Further, let curves $B(\lambda)$ and $Y(\lambda)$ approach one another gradually to become one curve named $W(\lambda)$. Then we would see the fields representing blue and yellow disappear gradually and only the black and white fields remain. Now, we can imagine that color vision was evolving in the opposite procedure. First, there was only one kind of visual cones in the human retina and only two totally different colors (black and white) could be discerned. Then, with color vision evolving, the cones split into two kinds that had different spectral sensitivities so that blue and yellow were also perceived. After that, the cones split into three kinds so that more colors were perceived.

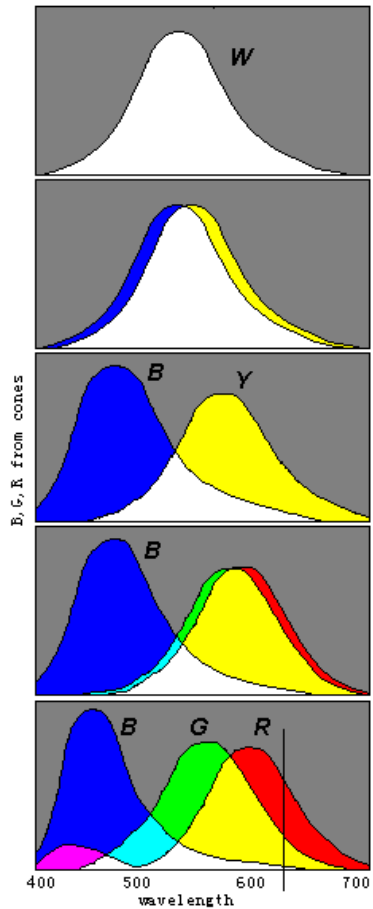

Figure 8. Evolution of color vision illustrated by splitting sensitivity curves

We may conclude that $n$ different kinds of cones can produce $2^n$ totally different color perceptions for $n$=1, 2, 3. As $n$=4, the conclusion seems also true. We have built a symmetrical model of four primary colors for robots (Lu, 1987). The model has 14 "unique colors", which can be symmetrically put on the surface of a ball, besides "white" (1, 1, 1, 1) and "black" (0, 0, 0, 0). We can get a "color" ball that has many properties very similar to those in the Newton color wheel.

The evolution of color vision might have come through somewhat different way. For example (see Fig. 9, deuteranopia-2), the curve $W(\lambda)$ first split into $R(\lambda)$ and $C(\lambda)$ related to cyan, instead of $B(\lambda)$ and $Y(\lambda)$, then $C(\lambda)$ split into $B(\lambda)$ and $G(\lambda)$.

## 7. **Color Blindness**

Color blindness has been discussed by many researchers (Hsia at el., 1965). It can also be easily explained by the sensitivity curves of cones that are too close each other. For example, monochromatism can be explained by the assumption that the sensitivity curves $B(\lambda)$, $G(\lambda)$ and $R(\lambda)$ have not yet separated from one curve; Red-green blindness can be explained by the assumption that the curves $G(\lambda)$ and $R(\lambda)$ have not separated yet.

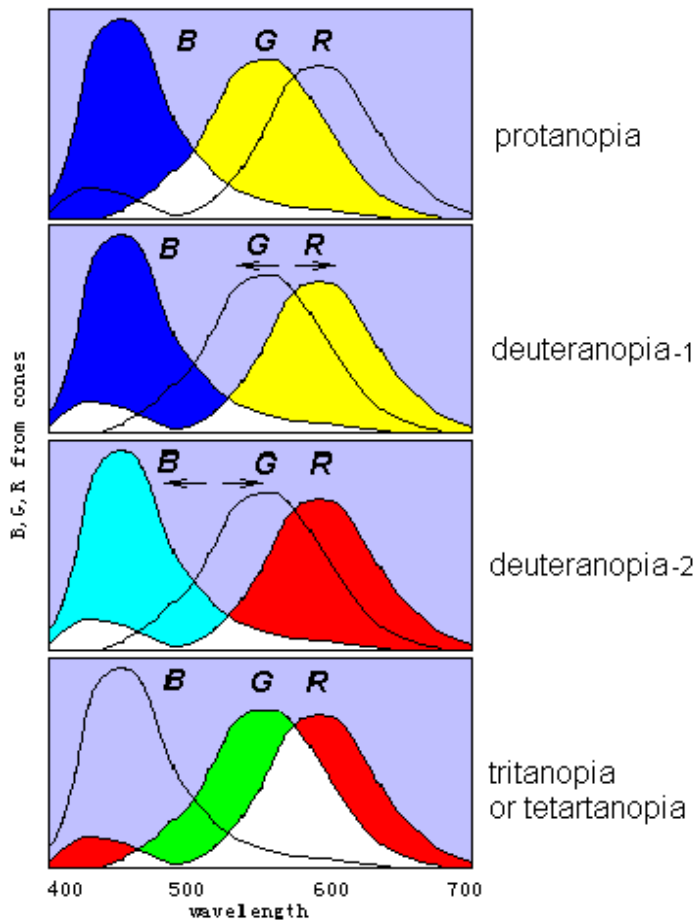

Figure 9 Different kinds of color blindness illustrated by incomplete separations of three sensitive curves

According to the decoding model, some red-green blindness can be identified as protanopia or deuteranopia only because the peak of $Y(\lambda)$ has shorter or longer wavelength. Tritanopia and tetartanopia can be illustrated by the assumption that the $B(\lambda)$ and $G(\lambda)$ ( or $B(\lambda)$ and $R(\lambda)$ ) have not separated yet so that each kind of color blindness can only perceive two chromatic colors: red and cyan (or green and magenta). All kinds of color blindness above can be imitated by the physical model with two of three of the 3DUs always obtain the same light inputs.

Assuming curve $G(\lambda)$ is split from the right curve or the left curve, we will have different deuteranopia: deuteranopia-1 and deuteranopia-2, which produce totally different color perceptions. However, according to philosophical analyses about the inverted spectrum, two kinds of color blindness must be equivalent and cannot be distinguished (Shoemaker, 1982; Lu, 1989).

Color anomalous can be explained in similar way.

## 8. Discussion

There are many reasons that make the decoding model convincible:
1) The model is concise, symmetrical, and without modification parameters.
2) It can more pictorially explain opponent process, color evolution, and color blindness.
3) In the popular zone model, adding red and green at the zone of visual cones forms yellow; yet,

adding red and green at the zone of visual nerves forms white. So, meanings of "red" and "green" in popular zone model are inconsistent. Yet, the decoding model has no this problem.

3) The decoding model is more compatible with the laws of color mixture that are used for color TV and computer graphics.

4) We can also use the decoding model to explain the phenomenon of negative after-image conveniently. For example, when the sensitivity of the $R$-cone falls down, $[BG\overline{R}]$ for cyanness will be over zero for a white color (1,1,1) so that white color looks cyan.

There is seemingly also a problem with the decoding model. In the popular zone model, there are only two pairs, instead of three pairs, of opponent colors. Seemingly psychological and physiological experiments support the affirmation that only two pairs of opponent colors exist. But, I think that the "red and green" in popular zone model is actually a pair of opponent colors between red-cyan and green-magenta. More than four unique colors were also affirmed by others (Hardin, C. L., 1985).

According to the decoding model, we can make two predictions. One is that there should be some fuzzy logic gates, which execute the operations of maximum, minimum, and even medium, in the human retina. Another is that there should be some chromatic opponent units in visual nerves, whose response curves have a horizontal line, instead of a neutral point, between positive and negative parts (see the right part of figure 5). These logic gates and opponent units have not been mentioned yet (which is another problem with the decoding model) either because most experiments were made with animals whose color vision is less complete than the man's, or because the guidance from appropriate theory was absent. For example, a widely used method for identifying a chromatic opponent unit is to find its neutral point (Volois and others, 1966); however, this method is not suitable for identifying the opponent unit suggested above. We believe that the predicted logic gates and the opponent units will be discovered soon by physiologists who pay attention to them.

## References

De Monasterio, F. M., Gouras P. and Tollhurst D J., (1975) Trichromatic color opponency in ganglion cells of therhesus monkey retina, J. Physiol. 252, 197-216

De Valois, Russell L.; Abramov, Israel; and Jacobs, Gerald H. (1966), "Analysis of response patterns of LGN cells", Journal of the Optical Society of America, 56(7): 966-977

Hardin, C. L. (1985) The resemblances of colors, Philosophical Studies 48, 35–47

Hsia, Y. and Grapham, C. H., (1965) Color blindness, in Visual Perception , C. H. Grapham ed., Wiley, New York, Chap. 14, Tab.14.2

Hunt, R. W. G., (1982) A model of color vision for predicting color appearance, Color Res. Appl. 7, 95-112

Hurvich, L. M. and Jameson D., (1957) An opponent-process theory of color vision, Psychological Review 64, 384-404

James D. Foley, Andries Van Dam, (1984) Title Fundamentals of Interactive Computer Graphics, Addison-Wesley

Judd, D. B., (1949) Response functions for types of vision according to the Muller theory, J. Res. Natl. Bur. Std. 42

Kaplan, E. and Shapley, (1982) R. M., X and Y cells in the lateral geniculale nucleus of Macaque monkeys, J. Physiol. 330, 125-143

Lu, C., (1986) New Theory of color vision and simulation of mechanism, Developments in Psychology (in China), 14, 36-45

Lu, C., (1987) Models of color vision for robots. Robot( A Journal of Chinese Society of Automation) , 1, No.6,39-46.

Lu, C., (1989) Decoding model and its verification, ACTA OPTICA SINICA, 9, 158-163

Lu, C., (1989a) Clarifying ostensible definition by the logical possibility of inverted spectrum, Modern Philosophy, No.2, 1989

Lu, C., (1991) B-Fuzzy set algebra and generalized mutual information formula, Fuzzy Systems and Mathematics, Vol. 5, No. 1, 76-80

Lu, C., (1999) A generalization of Shannon's information theory, Int. J. of General Systems, 28: (6) 453-490

Shoemaker S. (1982) The inverted spectrum，J．of Philosophy 79, 375-381

Smith, A. R.,(1978) Color Gamut Transform Pairs, Computer Graphics, Vol 12, No 3, 12-19

Walraven, P. L., (1961) On the Bezold-Brucke phenomenon, J. Opt. Soc. Am. 51, 1113-1116